\documentclass{article}
\usepackage{amssymb}
\usepackage{graphicx}
\usepackage{amsmath}
\usepackage{siunitx}
\usepackage{subcaption}
\usepackage{booktabs}
\usepackage{dsfont}

\usepackage[final]{corl_2025} 

\title{Learning More With Less: Sample-Efficient Model-Based RL for Loco-Manipulation}

\author{
  Benjamin Hoffman\\
  ETH Zürich, Switzerland\\
  \texttt{bhoffman@ethz.ch} \\
  \And
  Jin Cheng \\
  ETH Zürich, Switzerland\\
  \texttt{jin.cheng@inf.ethz.ch} \\
  \AND
  Chenhao Li \\
  ETH Zürich, Switzerland\\
  \texttt{chenhli@ethz.ch} \\
  \And
  Stelian Coros \\
  ETH Zürich, Switzerland\\
  \texttt{scoros@inf.ethz.ch} \\
}

\usepackage{cleveref}

\begin{document}
\maketitle


\begin{abstract}
By combining the agility of legged locomotion with the capabilities of manipulation, loco-manipulation platforms have the potential to perform complex tasks in real-world applications. To this end, state-of-the-art quadrupeds with manipulators, such as the Boston Dynamics Spot, have emerged to provide a capable and robust platform. However, the complexity of loco-manipulation control, as well as the black-box nature of commercial platforms, pose challenges for deriving accurate dynamics models and robust control policies. To address these challenges, we turn to model-based reinforcement learning (RL). We develop a hand-crafted kinematic model of a quadruped-with-arm platform which – employing recent advances in Bayesian Neural Network (BNN)–based learning – we use as a physical prior to efficiently learn an accurate dynamics model from limited data. We then leverage our learned model to derive control policies for loco-manipulation via RL. We demonstrate the effectiveness of our approach on state-of-the-art hardware using the Boston Dynamics Spot, accurately performing dynamic end-effector trajectory tracking even in low data regimes. Project website and videos: \href{https://sites.google.com/view/learning-more-with-less/}{sites.google.com/view/learning-more-with-less}.
\end{abstract}

\keywords{Sample Efficient Model-Based RL, Loco-Manipulation} 


\section{Introduction}  \label{introduction}
Legged robots have demonstrated impressive capabilities in navigating complex terrains, offering agility and adaptability as a result of continuous research efforts \cite{zhuang_robot_2023, jenelten_dtc_2024, li_learning_nodate, cheng_extreme_2024, hoeller_anymal_2024, choi_learning_2023}. However, while many systems excel at locomotion, their ability to interact with their environment remains limited. The integration of a manipulator onto a legged platform, i.e., legged loco-manipulation, holds the potential to bridge this gap. This combination enables a robot to both navigate challenging terrain and perform advanced manipulation tasks such as opening doors \cite{bellicoso_alma_2019}, grasping objects \cite{zimmermann_go_2021,liu_visual_2024, fu_deep_2022}, or possibly even interacting with objects in a dynamic setting such as catching or throwing a ball. State-of-the-art commercial robots such as the Boston Dynamics Spot quadruped, now equipped with an arm, have emerged to provide a capable and robust platform to perform such tasks \cite{zimmermann_go_2021}.

However, their proprietary, black-box nature complicates the development of an accurate dynamics model necessary to derive new control policies \cite{zimmermann_go_2021, li_fld_2024}. Simplified or hand-crafted modeling approaches alone often fall short in face of unknown internal controller behavior and complex dynamics, while purely model-free learning can demand large amounts of real-world data. Further, realizing robust loco-manipulation introduces new challenges. The coupling between a dynamic, moving base and a mounted manipulator creates complex, high-dimensional dynamics that are difficult to capture and control using classical methods \cite{fu_deep_2022}. 

To address both the challenge of learning an accurate system model, as well as performing robust loco-manipulation control, we turn to model-based reinforcement learning (RL) \cite{MAL-086}. Leveraging recent progress in dynamics learning with Bayesian Neural Networks (BNNs), specifically {\scshape Sim}-FSVGD \cite{rothfuss_bridging_2024}, and by developing a hand-crafted kinematic model of our platform, we efficiently learn a dynamics model of Spot with a mounted arm from real-world data. We use {\scshape Sim}-FSVGD to incorporate our kinematic model as a low-fidelity physical prior during BNN-learning, allowing us to learn an accurate model at low data requirements. Inspired by the recent success of RL-based control for loco-manipulation \cite{huang_creating_2022,ma_combining_2022,fu_deep_2022,ji_dribblebot_2023}, we then leverage our learned dynamics model to derive control policies via RL that enable Spot to accurately perform loco-manipulation tasks \cite{li2025robotic}. 

In summary, our main contributions are: (\textit{i}) we allow for learning an accurate dynamics model for a complex, black-box quadruped-with-arm platform from limited real-world data by developing a hand-crafted kinematic model and employing it as a physical prior for efficient BNN-learning leveraging {\scshape Sim}-FSVGD, (\textit{ii}) we use the learned dynamics model to derive control policies for loco-manipulation via RL, and (\textit{iii}) we demonstrate the effectiveness of our approach on the Boston Dynamics Spot with a manipulator, achieving improved dynamic end-effector trajectory tracking accuracy even at reduced data requirements compared to baseline methods. 


\begin{figure}
    \centering
    \includegraphics[width=1.0\linewidth]{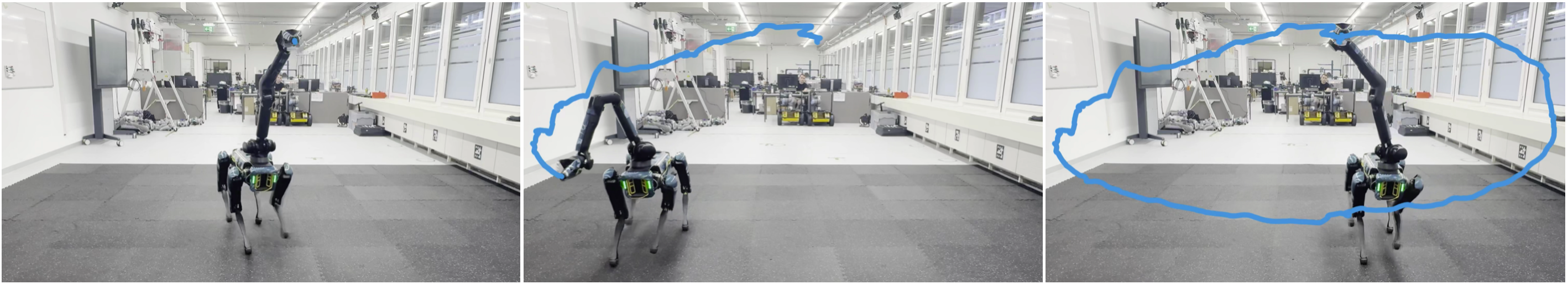}
    \caption{Boston Dynamics Spot in our experiments tracking an ellipsoidal reference trajectory.} 
    \label{fig:ellipse_teaser}
\end{figure}


\section{Related Work}  \label{related_work}

\paragraph{Loco-Manipulation Control}
Combining legged locomotion and manipulation to achieve dynamic mobile manipulation is an increasingly relevant problem that has been the focus of a considerable amount of research. We can generally make a distinction between platforms that use a robot's body \cite{sombolestan_hierarchical_2023, jeon_learning_2024} or legs \cite{huang_creating_2022, ji_dribblebot_2023, ji_hierarchical_2022} for manipulation, those that apply a hybrid approach \cite{lin_locoman_2024, whitman_generating_2017, forrai_event-based_2023}, and those that use a dedicated arm \cite{ma_combining_2022, fu_deep_2022, zimmermann_go_2021, liu_visual_2024, bellicoso_alma_2019, ferrolho_roloma_2023}. Especially this last category allows for combining the advances in legged locomotion with the benefits of a manipulator, enabling advanced tasks such as grasping stationary items \cite{fu_deep_2022, zimmermann_go_2021,liu_visual_2024}, opening doors \cite{bellicoso_alma_2019}, or wiping a whiteboard \cite{fu_deep_2022}.


However, due to the platform's complexity, loco-manipulation control is inherently a challenging, high-dimensional, and non-smooth control problem \cite{ma_combining_2022}. This calls for pursuing a closed-loop approach to allow for more robust and adaptive control instead of previous feed-forward trajectory optimization methods such as in \cite{zimmermann_go_2021}. To this end, RL has emerged as a powerful approach that enables robust legged locomotion in general \cite{choi_learning_2023, agarwal_legged_2022, margolis_walk_2022, miki_learning_2022, lee_learning_2020, hwangbo_learning_2019} and shows impressive performance in the setting of loco-manipulation \cite{huang_creating_2022,ma_combining_2022,fu_deep_2022,ji_dribblebot_2023,ji_hierarchical_2022,jeon_learning_2024}. 
While these prior works mainly focus on using \textit{model-free} RL, recently \textit{model-based} RL has shown potential for more sample-efficient robot learning across distinct locomotion and manipulation tasks \cite{wu2023daydreamer, Hafner2020Dream}. Motivated by this success, we leverage recent advances in \textit{model-based} RL and dynamics learning, primarily {\scshape Sim}-FSVGD \cite{rothfuss_bridging_2024}, to achieve accurate and robust loco-manipulation control, as well as better generalization to new tasks, even when data is scarce and a simulation environment is not available.

\paragraph{Modeling Robot Dynamics}
Although state-of-the-art quadrupeds like the Boston Dynamics Spot are very capable platforms, their usage in a research setting yields certain challenges. As a commercial product, knowledge of their built-in controller remains proprietary, and low-level control access is often restricted. However, an accurate system model is necessary to pursue a model-based or learning-based control method. To this end, \cite{zimmermann_go_2021} uses a simple parametrized dynamics model and parameter identification to simulate the platform's behavior. Yet, accurately modeling the dynamics of a complex system such as a quadruped with an attached arm is a challenging, dynamic, and high-DoF problem \cite{fu_deep_2022, li2025robotic}. In \cite{zimmermann_go_2021}, Spot fails to achieve the desired task in certain cases due to the model's inability to fully capture the robot's complex internal behavior. Especially the wrench and disturbances introduced by the arm are difficult to model accurately. 

In the face of these challenges, we pursue a more sophisticated approach. By developing a hand-crafted kinematic model of our system and employing methods for BNN-learning with physical priors, we incorporate domain knowledge into our learning pipeline to accurately capture the complex dynamics of our system in a sample-efficient manner. To this end, we leverage {\scshape Sim}-FSVGD \cite{rothfuss_bridging_2024}, a BNN-based method to learn accurate dynamics from limited data by incorporating our kinematic model as a low-fidelity physical prior during training. Further, by using \textit{Bayesian} Neural Networks, we tend to avoid the same overfitting behavior as non-bayesian methods \cite{rothfuss_bridging_2024}. This combined approach allows us to learn an accurate model of our system with improved sim-to-real performance, even in low-data regimes, which we can then leverage to derive robust control policies via RL.


\section{Preliminaries}  \label{background}

\subsection{Learning Robot Dynamics}
\paragraph{Learning with NNs}
We omit the preliminaries on (model-based) RL here and provide them in \Cref{mbrl}. In a robotics context, we want to use model-based RL to derive control policies $\pi$ that allow us to perform dynamic tasks, such as loco-manipulation. However, since we often lack a model of our system in practice, we first need to learn an accurate dynamics model of our robot. For this purpose, we consider a time-discretized dynamical system described as
\begin{equation}
    \mathbf{s}_{t+1} = f(\mathbf{s}_t, \mathbf{u}_t),
\end{equation}
where $\mathbf{s}_t \in \mathbb{R}^{n_s}$ is the state of the system at time $t$, $\mathbf{u}_t \in \mathbb{R}^{n_u}$ is the control input, and $f(\mathbf{s}_t, \mathbf{u}_t)$ is the unknown dynamics of the system. We now aim to learn a model $\hat{f}(\mathbf{s}_t, \mathbf{u}_t)$ that approximates the true dynamics of the system from a dataset of state-action-state transitions $(\mathbf{s}_t, \mathbf{u}_t, \mathbf{s}_{t+1})$. To this end, we can state our problem as learning an unknown dynamics function $\hat{f}: \mathcal{X} \rightarrow \mathcal{Y}$ from a dataset $\mathcal{D}=(\mathbf{X}^{\mathcal{D}}, \mathbf{y}^{\mathcal{D}})$ of size $N$, where our training inputs consist of the state-action pairs $\mathbf{X}^{\mathcal{D}}=\{\mathbf{x}_j\}_{j=1}^{N}$ and the target outputs $\mathbf{y}^{\mathcal{D}}=\{\mathbf{y}_j\}_{j=1}^{N}$ are the measured noisy observations of our dynamics, i.e., $\mathbf{y}_{j} = f(\mathbf{x}_j) + \epsilon_{j}$. We assume the noise $\epsilon$ to be i.i.d. and Gaussian with variance $\sigma^2$. Using a Neural Network (NN) to model $\hat{f}$, we can then formulate our learning problem as fitting a NN model $h_{\theta}: \mathcal{X} \rightarrow \mathcal{Y}$ with network weights $\theta$ from $\mathcal{D}$. We can use $h_{\theta}$ to define the conditional predictive distribution of our observations as $p(\mathbf{y}|\mathbf{x},\theta) = \mathcal{N}(\mathbf{y}|h_{\theta}(\mathbf{x}), \sigma^2)$.

\paragraph{Learning with BNNs}
This can be extended to learning a BNN by considering not only a single set of weights but a distribution over $\theta$. BNNs then infer a posterior distribution over the weights $p(\theta|\mathbf{X}^{\mathcal{D}}, \mathbf{y}^{\mathcal{D}}) \propto p(\mathbf{y}^{\mathcal{D}}|\mathbf{X}^{\mathcal{D}}, \theta)p(\theta)$ given the data-likelihood $p(\mathbf{y}^{\mathcal{D}}|\mathbf{X}^{\mathcal{D}}, \theta)$ and a known prior distribution $p(\theta)$. Under the assumption that, given $\theta$, each data point is conditionally independent, the likelihood can be factorized as $p(\mathbf{y}^{\mathcal{D}}|\mathbf{X}^{\mathcal{D}}, \theta) = \prod_{j=1}^{N} p(\mathbf{y}_j|\mathbf{x}_j, \theta)$. Finally, the predictive distribution for a new input $\mathbf{x}^*$ can be defined by marginalizing over the weights $\theta$ \cite{rothfuss_bridging_2024} as $p(\mathbf{y}^*|\mathbf{x}^*, \mathbf{X}^{\mathcal{D}}, \mathbf{y}^{\mathcal{D}}) = \int p(\mathbf{y}^*|\mathbf{x}^*, \theta) p(\theta|\mathbf{X}^{\mathcal{D}}, \mathbf{y}^{\mathcal{D}}) d\theta = \mathbb{E}_{\theta}[p(\mathbf{y}^*|\mathbf{x}^*, \theta) | \mathbf{X}^{\mathcal{D}}, \mathbf{y}^{\mathcal{D}}]$.

\subsection{Function Space Inference and Functional Priors}
\paragraph{FSVGD}
Performing posterior inference with BNNs, however, is challenging. Both the high-dimensionality of the weight space, as well as the over-parametrization of the mapping between $\theta$ and a likelihood function $p(\mathbf{y}^{\mathcal{D}}|\mathbf{X}^{\mathcal{D}}, \theta)$ render inference difficult \cite{rothfuss_bridging_2024,wang_function_2019}. The functional Stein Variational Gradient Descent (FSVGD) method \cite{wang_function_2019} addresses these issues by performing BNN inference in the space of regression functions $h: \mathcal{X} \rightarrow \mathcal{Y}$, rather than in the weight space of $\theta$. In function space, the posterior is formulated as $p(h|\mathbf{X}^{\mathcal{D}}, \mathbf{y}^{\mathcal{D}}) \propto p(\mathbf{y}^{\mathcal{D}}|\mathbf{X}^{\mathcal{D}},h)p(h)$,
where $p(h)$ is a stochastic prior distribution over $h: \mathcal{X} \rightarrow \mathcal{Y}$ with index space $\mathcal{X}$ and value space $\mathcal{Y}$ \cite{sun_functional_2019}. This allows for the functional inference to be restated in a tractable form by using finite measurement sets $\mathbf{X} := [\mathbf{x}_1, ..., \mathbf{x}_k] \in \mathcal{X}^k, k \in \mathbb{N}$ that allow for characterizing a stochastic process by marginals of function values $\rho(\mathbf{h}^{\mathbf{X}}) := \rho(h(\mathbf{x}_1), ..., h(\mathbf{x}_k))$ and subsequently stating the functional posterior as $p(\mathbf{h}^{\mathbf{X}}|\mathbf{X},\mathbf{X}^{\mathcal{D}}, \mathbf{y}^{\mathcal{D}}) \propto p(\mathbf{y}^{\mathcal{D}}|\mathbf{X}^{\mathcal{D}}, \mathbf{h}^{\mathbf{X}}) p(\mathbf{h}^{\mathbf{X}})$, for the measurement sets $\mathbf{X}$ \cite{oksendal_stochastic_2003}. FSVGD approximates this posterior by maintaining $L$ parameter particles ${\theta_1, ..., \theta_L}$ and iteratively re-sampling $\mathbf{X}$ as a random subset of $\mathcal{X}$ with $\mathbf{X} \sim \mu$ where $\mu$ is an arbitrary distribution supported on $\mathcal{X}$. FSVGD then updates the particles using 
\begin{equation}\label{eq:FSVGD_update}
    \theta_l = \theta_l - \gamma J_l u_l.
\end{equation}
Here $J_l = (\nabla_{\theta_l} \mathbf{h}_{\theta_l}^{\mathbf{X}})^{\top}$ is the NN Jacobian, $u_l = \frac{1}{L} \sum_{i=1}^L \mathbf{K}_{li} \nabla_{\mathbf{h}_{\theta_i}^{\mathbf{x}}} \ln p(\mathbf{h}_{\theta_l}^{\mathbf{x}} | \mathbf{X}, \mathbf{X}^D, \mathbf{y}^D) + \nabla_{\mathbf{h}_{\theta_l}^{\mathbf{x}}} \mathbf{K}_{li}$ is the SVGD update \cite{liu_stein_2019} in function space and $\mathbf{K} = [k(\mathbf{h}_{\theta_l}^{\mathbf{x}}, \mathbf{h}_{\theta_i}^{\mathbf{x}})]_{li}$ is the gram matrix, based on a kernel function $k$, between the measurement points and the function values \cite{rothfuss_bridging_2024,wang_function_2019}.

\paragraph{{\scshape Sim}-FSVGD}
The {\scshape Sim}-FSVGD method \cite{rothfuss_bridging_2024} extends FSVGD by using an \textit{informed} functional prior $p(h)$ for a function $h: \mathcal{X} \rightarrow \mathcal{Y}$ that incorporates both a \textit{domain-model process} and a \textit{sim-to-real prior}. {\scshape Sim}-FSVGD factorizes the prior over the output dimensions as $p(h) = \prod_{i=1}^{n_s} p(h_i)$, treating each $h_i: \mathbf{X} \rightarrow \mathbb{R}$ as an independent function. The \textit{domain-model process} allows for integrating prior domain knowledge of the system via a low-fidelity simulation model $g(\mathbf{x}, \phi)$, e.g., derived from first-principle physics, where $\phi$ are the model parameters. As the exact model parameters are unknown, we can randomly sample them from a plausible range as $\phi \sim p(\phi)$ and create distinct simulation models per parameter set, implicitly creating a stochastic process of functions. The \textit{sim-to-real prior} addresses the gap between a simulation model and the actual system dynamics $f(\mathbf{x})$ by adding a sim-to-real gap process as a Gaussian Process (GP) $p(\tilde{h}_i)$ per output dimension $i = 1, ..., n_s$. {\scshape Sim}-FSVGD uses a zero-mean GP with isotropic kernel $k(\mathbf{x}, \mathbf{x}^{\prime}) = \nu^2 \rho(||\mathbf{x} - \mathbf{x}^{\prime}||/l)$, where the lengthscale $l$ and variance $\nu^2$ are hyperparameters that allow us to incorporate our assumptions about the actual sim-to-real gap.

The combined stochastic process prior $p(h)$ is then defined implicitly via the marginal distributions implied by independently sampling conditional random vectors from each process and adding them: $\mathbf{h}_i^{\mathbf{x}} = \left[ g_i(\mathbf{x}_1, \phi), \dots, g_i(\mathbf{x}_k, \phi) \right]^{\top} + \tilde{\mathbf{h}}_i^{\mathbf{x}}$, where $\phi \sim p(\phi)$ and $\tilde{\mathbf{h}}^{\mathbf{X}}_i \sim \mathcal{N}(\tilde{\mathbf{h}}^{\mathbf{X}}_i|0, \mathbf{K})$ \cite{rothfuss_bridging_2024}.
Lastly, {\scshape Sim}-FSVGD relies on the same update rule as in \Cref{eq:FSVGD_update} to update the particles $\theta_l$. The stochastic process prior score $\nabla_{\mathbf{h}^{\mathbf{X}}} \ln p(\mathbf{h}^{\mathbf{X}})=\sum_{i}^{n_s}\nabla_{\mathbf{h}^{\mathbf{X}}_i} \ln p(\mathbf{h}^{\mathbf{X}}_i)$ is approximated using a Gaussian approximation of the prior process, i.e., $p(\mathbf{h}_i^{\mathbf{X}}) \sim \mathcal{N}(\mu_i^{\mathbf{X}}, \Sigma_i^{\mathbf{X}})$. The approximation is constructed by sampling the measurement set $\mathbf{X}$ from a distribution $\mu$ supported on $\mathcal{X}$, sampling $m=1, ..., P$ vectors of function values $\mathbf{h}_{i,m}^{\mathbf{X}} \sim p(\mathbf{h}_i^\mathbf{X})$ and computing their mean $\mu_i^{\mathbf{X}}$ and covariance $\Sigma_i^{\mathbf{X}}$.


\section{Learning Control Policies for Loco-Manipulation} \label{methods}
Our work focuses on learning an accurate dynamics model of our robot from limited data to subsequently derive control policies for loco-manipulation tasks via RL. To this end, we develop a hand-crafted kinematic model $\hat{f}_{kin}$ of our quadruped-with-arm platform. We then use $\hat{f}_{kin}$ as a physical prior and leverage {\scshape Sim}-FSVGD to efficiently learn a model $\hat{f}$ from data that approximates our platform's true dynamics $f$. Subsequently, we develop a reward structure ${r}$ and use our learned model $\hat{f}$ to derive a control policy $\pi$ for end-effector trajectory tracking using Soft-Actor-Critic (SAC) \cite{haarnoja_soft_2018}. In this section, we present our control approach, develop our kinematic model $\hat{f}_{kin}$, show how we use it and {\scshape Sim}-FSVGD to learn $\hat{f}$, and finally discuss our policy learning process.

\subsection{Robot State and Control Input}
We define our base state as the position, orientation and velocity of the robot's base on a 2D plane in world frame $W$, i.e., $\mathbf{p}^{\text{base}} = [x^{\text{base}}, y^{\text{base}}, \theta^{\text{base}}]$ and $\mathbf{v}^{\text{base}} = [v_{x}^{\text{base}}, v_{y}^{\text{base}}, \omega^{\text{base}}]$. Here $\theta^{\text{base}}$ is the yaw angle of the robot's base, and $\omega^{\text{base}}$ is the corresponding angular velocity. For the end-effector, we represent the state using the 3D position and velocity in world frame $W$, i.e., $\mathbf{p}^{\text{ee}} = [x^{\text{ee}}, y^{\text{ee}}, z^{\text{ee}}]$ and $\mathbf{v}^{\text{ee}} = [v_{x}^{\text{ee}}, v_{y}^{\text{ee}}, v_{z}^{\text{ee}}]$. Stacking the individual components, our state vector becomes $\mathbf{s}= \begin{bmatrix}\mathbf{p}^{\text{base}}, \mathbf{v}^{\text{base}}, \mathbf{p}^{\text{ee}}, \mathbf{v}^{\text{ee}} \end{bmatrix} \in \mathbb{R}^{12}$. We provide an overview of the respective frames in \Cref{frames}.

To control our robot, we use velocity commands. For the base of the robot, we command the $v_{x}^{\text{base}}$, $v_{y}^{\text{base}}$ and $\omega^{\text{base}}$ velocities on a 2D plane, yielding the control input $\mathbf{u}^{\text{base}} = [u_{vx}^{\text{base}}, u_{vy}^{\text{base}}, u_{\omega}^{\text{base}}]$. We express the velocities in the body frame $B$ and apply them at the robot's center of mass. For the end-effector, we command the $v_{x}^{\text{ee}}$, $v_{y}^{\text{ee}}$ and $v_{z}^{\text{ee}}$ velocities in 3D, i.e., $\mathbf{u}^{\text{ee}} = [u_{vx}^{\text{ee}}, u_{vy}^{\text{ee}}, u_{vz}^{\text{ee}}]$. Again, the velocities are expressed in the body frame $B$ and are applied at the center of the end-effector frame $H$, which sits at the center of the gripper. Our control input is then given as $\mathbf{u} = \begin{bmatrix}\mathbf{u}^{\text{base}}, \mathbf{u}^{\text{ee}}\end{bmatrix} \in \mathbb{R}^{6}.$



\subsection{Developing a Dynamics Model}
We model the dynamics of our robot as the time discretized system $\mathbf{s}_{t+1} = \hat{f}(\mathbf{s}_t, \mathbf{u}_t)$, where $\mathbf{s}_t \in \mathbb{R}^{n_s}$ is the state at time $t$ with $n_s=12$, $\mathbf{u}_t \in \mathbb{R}^{n_u}$ is the control input with $n_u = 6$, and $\hat{f}(\mathbf{s}_t, \mathbf{u}_t)$ are the approximated dynamics. We now develop our dynamics model in two steps. First, we create a hand-crafted kinematic model $\hat{f}_{kin}$ of our robot derived from first principles using domain knowledge. In a second step, we use our model as a physical prior, i.e., to create the \textit{domain-model process} within {\scshape Sim}-FSVGD, and efficiently learn a dynamics model $\hat{f}$ from real-world data.

\paragraph{Kinematic Model}
We derive our kinematic model from first principles and use the Forward Euler Method with time step $\Delta t$ to propagate our state. To better capture the complex dynamics of our platform, we enhance our equations with the parameters $\mathbf{\alpha} \in \mathbb{R}^{6 \times 1}$, $\mathbf{\beta} \in \mathbb{R}^{12 \times 1}$, and $\mathbf{\gamma} \in \mathbb{R}^{6 \times 1}$. As our actions $\mathbf{u}_t$ are given in the body frame $B$ and our state $\mathbf{s}_t$ is in the world frame $W$, we first convert our inputs to $\mathbf{u}_t^W$ using the base's current yaw angle $\theta^{\text{base}}_t$, i.e., 


\begin{equation}
\mathbf u_t^{W}
  = \begin{bmatrix}
      R(\theta_t^{\text{base}})\,\mathbf u_t^{\text{base},B} \\[4pt]
      R(\theta_t^{\text{base}})\,\mathbf u_t^{\text{ee},B}
    \end{bmatrix},
\quad \text{where} \quad
R(\theta) \;:=\;
\begin{bmatrix}
  \cos(\theta) & -\sin(\theta) & 0\\
  \sin(\theta) &  \cos(\theta) & 0\\
  0          &  0          & 1
\end{bmatrix}.
\label{eq:action_conversion}
\end{equation}

We can then derive the update equations for the base velocity $\mathbf{v}^{\text{base}}$ and position $\mathbf{p}^{\text{base}}$, using $\mathbf{A}_{\mathrm{base}} := \operatorname{diag}(\boldsymbol{\alpha}_{1:3})$ and $\mathbf{\Gamma}_{\mathrm{base}} := \operatorname{diag}(\boldsymbol{\gamma}_{1:3})$, as



\begin{equation}
\label{eq:base_updates}
\begin{aligned}
\mathbf{v}_{t+1}^{\text{base}} &=
    \mathbf{A}_{\mathrm{base}}\,\mathbf{v}_{t}^{\text{base}}
    +(\mathbf{I}_3-\mathbf{A}_{\mathrm{base}}\,)
      \mathbf{u}_{t}^{\text{base},W}
    +\boldsymbol{\beta}_{1:3},\\[2pt]
\mathbf{p}_{t+1}^{\text{base}} &=
    \mathbf{p}_{t}^{\text{base}}
    +\Delta t\,\mathbf{\Gamma}_{\mathrm{base}}\,
      \mathbf{v}_{t+1}^{\text{base}}
    +\boldsymbol{\beta}_{4:6}.
\end{aligned}
\end{equation}

For the end-effector updates, we need to consider the base's movements in addition to the end-effector velocity commands. Incorporating the horizontal linear velocities of the base $v_{x,t}^{\text{base}}$ and $v_{y,t}^{\text{base}}$ follows simply via addition. However, to consider the base's angular velocity $\omega_{t}^{\text{base}}$ in the end-effector's movement, we first need to calculate the induced velocity on the end-effector as $\mathbf{v}^{\text{ind}} =
\begin{bmatrix}
-\omega_{t+1}^{\text{base}} \cdot d \cdot \sin(\phi),
\omega_{t+1}^{\text{base}} \cdot d \cdot \cos(\phi),
0
\end{bmatrix} \label{eq:induced_vel}$. Here $d=\sqrt{ \left( x_{t}^{\text{ee}} - x_{t}^{\text{base}} \right)^2 + \left( y_{t}^{\text{ee}} - y_{t}^{\text{base}} \right)^2 }$ is the distance between the base's rotational axis and the end-effector, and $\phi = \mathrm{arctan2}\left( y_{t}^{\text{ee}} - y_{t}^{\text{base}}, \, x_{t}^{\text{ee}} - x_{t}^{\text{base}} \right)$ is the angle of the induced velocity vector in the global frame $W$. We can then update the end-effector velocity $\mathbf{v}^{\text{ee}}$ and position $\mathbf{p}^{\text{ee}}$, again using $\mathbf{A}_{\mathrm{ee}} := \operatorname{diag}(\boldsymbol{\alpha}_{4:6})$, $\mathbf{\Gamma}_{\mathrm{ee}} := \operatorname{diag}(\boldsymbol{\gamma}_{4:6})$ and $\mathbf{D}_{ee} := \operatorname{diag}(1,1,0)$, as


\begin{equation}
\label{eq:ee_updates_compact}
\begin{aligned}
\mathbf{v}_{t+1}^{\text{ee}} &=
      \mathbf{A}_{\mathrm{ee}}\,\mathbf{v}_{t}^{\text{ee}}
    + (\mathbf{I}_3-\mathbf{A}_{\mathrm{ee}})\,\mathbf{u}_{t}^{\text{ee},W}
    + \boldsymbol{\beta}_{7:9}
    + \mathbf{D}_{ee}\,\mathbf{v}_{t+1}^{\text{base}}
    + \mathbf{v}^{\text{ind}},\\[2pt]
\mathbf{p}_{t+1}^{\text{ee}} &=
      \mathbf{p}_{t}^{\text{ee}}
    + \Delta t\,\mathbf{\Gamma}_{\mathrm{ee}}\,\mathbf{v}_{t+1}^{\text{ee}}
    + \boldsymbol{\beta}_{10:12}.
\end{aligned}
\end{equation}

\paragraph{BNN Model}
To learn a BNN model of our robot from data, we leverage {\scshape Sim}-FSVGD \cite{rothfuss_bridging_2024}, as detailed in section \ref{background}. {\scshape Sim}-FSVGD allows us to incorporate our prior knowledge by using our kinematic model $\hat{f}_{kin}$ to create the \textit{domain-model process}, where our system parameters $\phi$ are the set of parameters used in our update equations, i.e., $\phi = [\mathbf{\alpha}, \mathbf{\beta}, \mathbf{\gamma}] \in \mathbb{R}^{24 \times 1}$. Note that, as {\scshape Sim}-FSVGD randomly samples parameter sets as $\phi \sim p(\phi)$ to implicitly create a stochastic process of functions, we do \textit{not} fit the parameters of our kinematic model from data beforehand. However, we use real-world data to heuristically estimate a plausible range for our parameters. We then use a similar \textit{sim-to-real prior} as \cite{rothfuss_bridging_2024} and learn our dynamics model $\hat{f}$ from real-world data.

\subsection{Policy Learning}
Having learned a model of our system, we now turn to deriving loco-manipulation control policies via RL. We employ SAC \cite{haarnoja_soft_2018} and condition our policy on end-effector goal positions $\mathbf{g}^{\text{ee}} = [x_{g}^{W}, y_{g}^{W}, z_{g}^{W}]$. To this end, we uniformly sample an initial state $\mathbf{s}_0$ and end-effector goal position $\mathbf{g}^{\text{ee}}$ at each episode and construct a goal conditioned state vector $\mathbf{s}_{\text{cond}} = [\mathbf{s}, \mathbf{g}^{\text{ee}}] \in \mathbb{R}^{15 \times 1}$, while our action space remains our control input $\mathbf{u} \in \mathbb{R}^{6 \times 1}$. To guide our policy learning, we design a reward structure $r(\mathbf{s}_{\text{cond}}, \mathbf{u})$ that encourages the end-effector to smoothly move towards the goal while keeping the end-effector within a physically valid range, consisting of a state-goal distance reward $r_{\text{state}}$, an end-effector to base distance reward $r_{\text{ee-base}}$ and a regularizing action reward $r_{\text{action}}$. 

\paragraph{\textit{$r_{\text{state}}$}}
To drive the end-effector towards a goal position, we assign a full reward when the distance $d_{\text{ee-goal}} = \lVert \mathbf{p}^{\text{ee}} - \mathbf{g}^{\text{ee}} \rVert$ between the end-effector position $\mathbf{p}^{\text{ee}}$ and the goal position $\mathbf{g}^{\text{ee}}$ lies within $(0,b)$. Outside these bounds, we smoothly decrease the reward using a long-tailed sigmoid function $\sigma_{m,a}(x)$ \footnote{We borrow the definition from \cite{tassa_dm_control_2020}, i.e., $\sigma_{m,a}(x) = \left( \left(x m^{-1} \sqrt{a^{-1} - 1}\right)^{2} + 1 \right)^{-1}$}, with a defined value $a$ at margin $m$, creating a smooth reward with infinite-support and range $[0,1]$: $r_{\text{state}}\!\left(\mathbf{s}_{\text{cond}}\right)
  \;=\;
  \mathds{1}_{\{\,d_{\text{ee-goal}}\le b\,\}}
  \;+\;
  \mathds{1}_{\{\,d_{\text{ee-goal}}> b\,\}}\,
  \sigma_{m,a}\!\left(d_{\text{ee-goal}}-b\right).
$


\paragraph{\textit{$r_{\text{ee-base}}$}}
The second component encourages the distance $d_{\text{ee-base}} = \lVert \mathbf{p}^{\text{ee}} - \mathbf{p}^{\text{base}} \rVert$ between the end-effector position $\mathbf{p}^{\text{ee}}$ and the base position $\mathbf{p}^{\text{base}}$ to stay within a physically valid range given by the arm's length $l_{\text{arm}}$. To achieve this, we use the same approach as above with the bounds $(0,l_{\text{arm}})$, resulting in: 
$r_{\text{ee-base}}\!\bigl(\mathbf{s}_{\text{cond}}\bigr)
  \;=\;
  \mathds{1}_{\{\,d_{\text{ee-base}}\le l_{\text{arm}}\,\}}
  \;+\;
  \mathds{1}_{\{\,d_{\text{ee-base}}> l_{\text{arm}}\,\}}\,
  \sigma_{m,a}\!\bigl(d_{\text{ee-base}}-l_{\text{arm}}\bigr)
$.


\paragraph{\textit{$r_{\text{action}}$}}
Lastly, we include an action cost term that penalizes inefficient policies. However, instead of weighting each control input equally, we encourage the use of end-effector movements over body movements by assigning a higher weight $\lambda_{\text{base}}$ to the base actions than the end-effector actions weight $\lambda_{\text{ee}}$: $r_{\text{action}}(\mathbf{u}) = -\left( \lambda_{\text{base}}\lVert \mathbf{u}^{\text{base}}\rVert^2 + \lambda_{\text{ee}}\lVert \mathbf{u}^{\text{ee}}\rVert^2 \right)$.

We calculate our final reward as a weighted sum of the three components, i.e., $r(\mathbf{s}_{\text{cond}}, \mathbf{u})= w_{1}r_{\text{state}}(\mathbf{s}_{\text{cond}}) + w_{2}r_{\text{ee-base}}(\mathbf{s}_{\text{cond}}) + w_{3}r_{\text{action}}(\mathbf{u})$, where $w_{1}$, $w_{2}$, and $w_{3}$ are tuned heuristically.

\section{Experimental Results}  \label{results}
In this section, we present our experiments and results, evaluating the effectiveness of our model-based RL approach at learning dynamic loco-manipulation control policies for a complex quadruped-with-arm platform in a data-efficient manner. To this end, we compare the performance of the dynamics model learned using {\scshape Sim}-FSVGD with our kinematic model as a physical prior, which we simply label {\scshape Sim}-FSVGD, to two baseline models, {\scshape Sim}-MODEL and FSVGD, across different training set sizes. We evaluate both the sim-to-real transfer performance of the models, as well as the real-world loco-manipulation performance of the control policies derived from them. 

In the following, we describe our baseline models, our experiment platform (the Boston Dynamics Spot), as well as our data collection and data processing steps. Subsequently, we introduce our three experiments: \textit{Model Validation} and our hardware experiments \textit{Ellipse Tracking} and \textit{Helix Tracking}. Finally, we present the results of our evaluation.

\subsection{Baseline Models}
We consider two baseline models, {\scshape Sim}-MODEL and FSVGD \cite{wang_function_2019}. {\scshape Sim}-MODEL is our hand-crafted kinematic model with the parameters $\mathbf{\alpha}$, $\mathbf{\beta}$, and $\mathbf{\gamma}$ fitted from real-world data using the optimizer \textit{Adam} \cite{kingma_adam_2017}. Notably, we use an \textit{unfitted} version of the same kinematic model as a low-fidelity physical prior in the {\scshape Sim}-FSVGD approach. FSVGD is a BNN-based method widely applied in deep learning and, contrary to {\scshape Sim}-FSVGD, FSVGD does \textit{not} use an informed prior.

\subsection{Experiment Setup}
\paragraph{Boston Dynamics Spot Quadruped}
For our experiments, we use the Boston Dynamics Spot, a state-of-the-art quadruped robot equipped with an arm for manipulation. We pass our control input and collect state measurements over Wi-Fi from a PC via Spot’s high-level Python SDK. Spot relies on an unkown state estimator that fuses data from onboard sensors and cameras, and an unkown onboard controller that executes our commands. \Cref{frames} shows Spot and its reference frames.


\paragraph{Data Collection and Processing}
We collect our training data by interacting with Spot, manually controlling its base $\mathbf{v}^{\text{base}}$ velocities using an Xbox controller and its end-effector $\mathbf{v}^{\text{ee}}$ velocities using a 3D Space Mouse. We send commands and collect data at \SI{15}{\Hz}. We collect transitions consisting of the current state $\mathbf{s}_t$, the commanded action $\mathbf{u}_t$, and the next state $\mathbf{s}_{t+1}$.  However, instead of using our dynamics model to predict the next system state directly, we predict the change in the state. To this end, we adapt our dataset by creating a new set where our input remains the state action pair $\mathbf{x}_t=[\mathbf{s}_t,\mathbf{u}_t]$ but our target output becomes the state difference $\mathbf{y}_t=[\mathbf{s}_{t+1} - \mathbf{s}_t]$. Additionally, we encode the base's yaw angle $\theta^{base}$ as $(\sin(\theta^{base}),\cos(\theta^{base}))$ to avoid discontinuities and provide a representation more suitable for NNs \cite{geist_learning_2024}. Further, our control setup has a delay (ca. two timesteps, i.e., ~\SI{133}{\ms}) between the command and execution of an action. To compensate for this, we append the previous two actions $[\mathbf{u}_{t-2}, \mathbf{u}_{t-1}]$ to the state $\mathbf{s}_{t}$. Our model input becomes $\mathbf{x}_t=[\mathbf{s}_{t},\mathbf{u}_{t-2}, \mathbf{u}_{t-1},\mathbf{u}_t] \in \mathbb{R}^{31 \times 1}$ while our target output is $\mathbf{y}_t=[\mathbf{s}_{t+1} - \mathbf{s}_t] \in \mathbb{R}^{13 \times 1}$. We then create a training set for supervised learning by sampling i.i.d. from the collected transitions.

\begin{figure}
    \centering
    \includegraphics[width=1.0\linewidth]{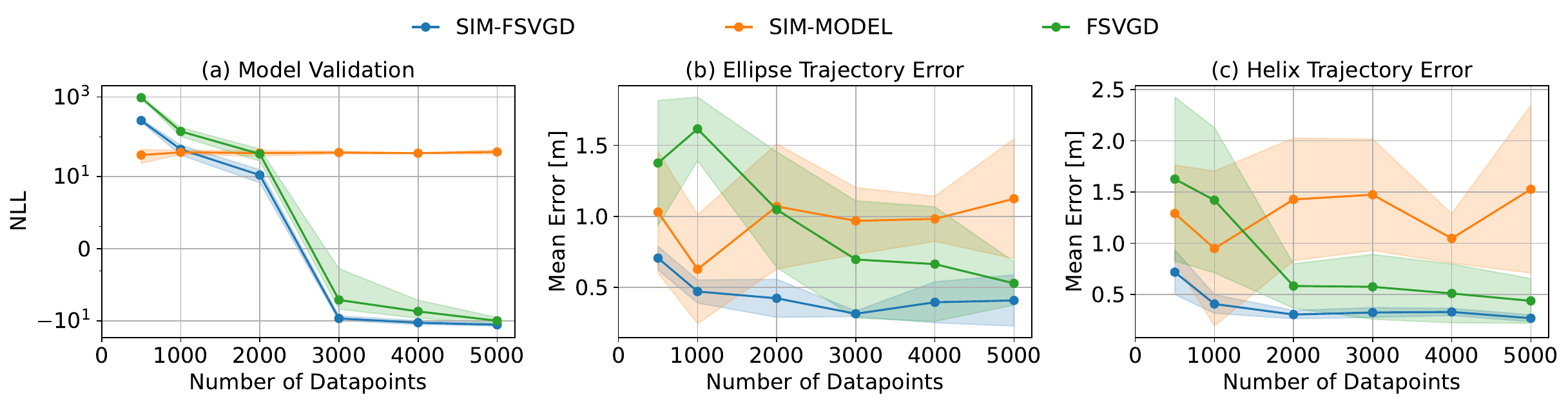}
    \caption{(a) We compare the dynamics models' sim-to-real transfer performance across increasing training set sizes by evaluating their NLL scores. {\scshape Sim}-FSVGD consistently outperforms the FSVGD baseline (especially in low data regimes) and the {\scshape Sim}-MODEL baseline from $N\geq2000$ onwards. \\ (b) We compare the mean error our policies achieve on the ellipsoidal goal trajectory across different training set sizes. The policies learned using the {\scshape Sim}-FSVGD model outperform those learned using our baseline models across all set sizes and especially at smaller training set sizes ($N<3000$). \\ (c) Similarly, for the helix trajectory, the policies learned using the {\scshape Sim}-FSVGD model outperform those learned using FSVGD and our kinematic {\scshape Sim}-MODEL across all training set sizes.}
    \label{fig:combined_errors}
\end{figure}

\begin{figure}
    \centering
    \vspace{0.2cm}
    \includegraphics[width=1.0\linewidth]{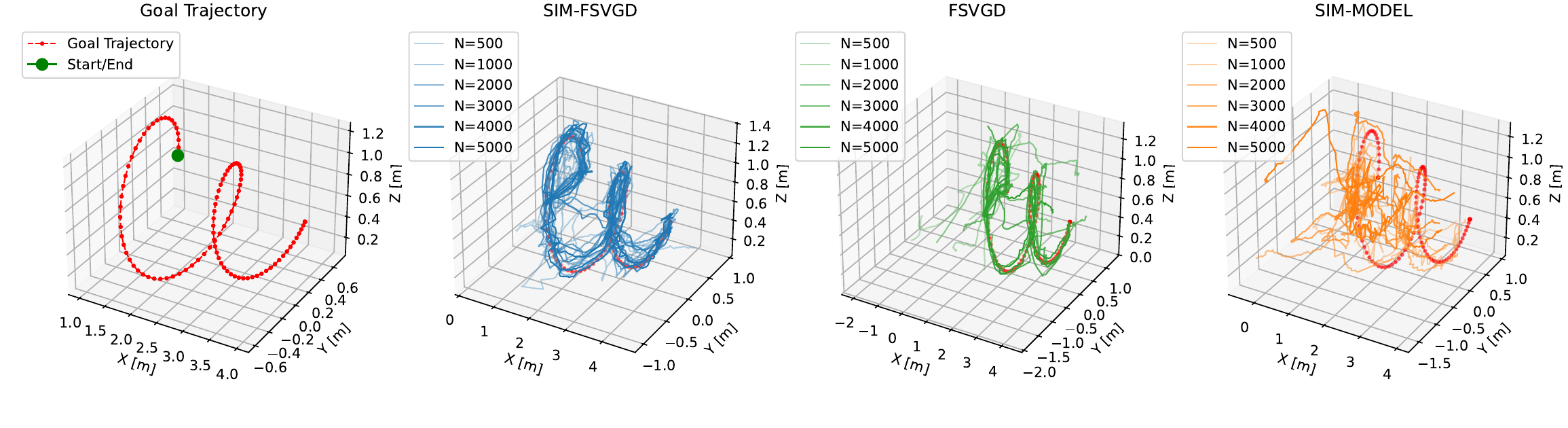}
    \caption{We plot the realized helix trajectories for all sample sizes and seeds. The policies learned using the {\scshape Sim}-FSVGD model follow the reference trajectory more closely than both baselines. Note that trajectories on course for collision were stopped early and we then plot only a truncated version.}
    \label{fig:helix_traj}
\end{figure}

\subsection{Results}

\paragraph{Model Validation}
Spot’s unknown internal controller behavior, along with the wrench and disturbances introduced by its arm, give rise to dynamics that are difficult to capture and model accurately. To evaluate how well our dynamics learning approach bridges this sim-to-real gap, we train the {\scshape Sim}-FSVGD and FSVGD models on a dataset sampled i.i.d. from our collected transitions and use the same dataset to fit the {\scshape Sim}-MODEL parameters. We evaluate their sim-to-real performance using the negative log-likelihood (NLL) scores they achieve on a real-world test set and compare our {\scshape Sim}-FSVGD model's performance to the baseline models across increasing training set sizes. We repeat our experiment with three random seeds and average the results.

We observe that {\scshape Sim}-FSVGD outperforms FSVGD across all training set sizes (\Cref{fig:combined_errors}). Especially in low data regimes, leveraging our physical prior via {\scshape Sim}-FSVGD helps us achieve NLL scores significantly lower than FSVGD. At $N=1000$, {\scshape Sim}-FSVGD performs similarly to our hand-crafted kinematic {\scshape Sim}-MODEL and surpasses it beyond $N\geq2000$. While the BNN-based models improve with growing training set sizes, our kinematic {\scshape Sim}-MODEL's NLL scores remain consistent throughout, which is expected as we fit $24$ parameters from an abundant amount of data.


\paragraph{Shape Tracking}
We now leverage our learned dynamics to derive loco-manipulation control policies via RL. To evaluate the effectiveness of our approach on hardware, we use the learned policies to dynamically track two shapes with Spot's end-effector: an ellipse and a helix. 
We show Spot in action in \Cref{fig:ellipse_teaser}. We compare the performance of the policies learned using our {\scshape Sim}-FSVGD dynamics model to the baselines across increasing training set sizes, repeating the experiment with three random seeds. We evaluate the performance via the mean error of the realized trajectories versus the reference trajectory: $\frac{1}{T}\sum_{t=1}^{T}||\mathbf{p}_t^\text{ee} - \mathbf{g}_t^{\text{ee}}||_2$, where $\mathbf{g}_t^{\text{ee}}$ is the end-effector goal at time $t$. 

The policies learned using our {\scshape Sim}-FSVGD model outperform both baselines across all training set sizes (\Cref{fig:combined_errors}). Especially in low-data regimes ($N\le2000$), leveraging our physical prior via {\scshape Sim}-FSVGD allows us to still learn an accurate dynamics model, enabling us to derive control policies that achieve significantly lower errors than both the FSVGD and the {\scshape Sim}-MODEL baselines (147\% and 153\% higher at $N=2000$ for the ellipse, respectively). While the performance of the policies learned using the FSVGD model improves with increasing training set sizes (as the model accuracy itself improves), even at $N=5000$, their error is still larger than that of the policies learned using the {\scshape Sim}-FSVGD model at $N=1000$. The plotted trajectories underline our results (\Cref{fig:helix_traj} and \Cref{realized_trajs}); the policies learned using the {\scshape Sim}-FSVGD model follow the reference trajectory more closely than both baseline methods. These hardware results demonstrate the effectiveness of our approach at efficiently learning loco-manipulation control policies for a complex platform.

\section{Conclusion}  \label{conclusion}
In this work, we address the problem of learning policies for loco-manipulation control on a quadruped platform with a manipulator. We develop a hand-crafted kinematic model which, by employing advances in dynamics learning with BNNs (i.e., {\scshape Sim}-FSVGD \cite{rothfuss_bridging_2024}), we leverage as a physical prior to efficiently learn a dynamics model of our system from limited data. In our hardware experiments, we use our learned dynamics model to derive loco-manipulation policies via RL, achieving improved dynamic end-effector trajectory tracking accuracy even at reduced data requirements compared to baseline methods. Our results demonstrate the effectiveness of our approach on a complex, commercial loco-manipulation system with a proprietary, black-box nature, such as Spot.

\paragraph{Limitations} Our approach shows certain shortcomings that could be addressed in future work. Although our tracked trajectories cover 3D space, they do not fully exploit the dynamic capabilities of the platform. Future work could investigate trajectories that require even faster motion of the base, such as catching a ball, or longer trajectories. Further, our current state and action space do not yet include the end-effector's orientation. Exploring how to incorporate these additional degrees of freedom into our model and control policies could be beneficial for performing more complex tasks.


\clearpage


\bibliography{example}  


\clearpage
\appendix
\section{Preliminaries on Model-Based Reinforcement Learning}\label{mbrl}
In this section, we provide a brief introduction to model-based RL in the context of robot control. In an RL setting, our problem of learning control policies can be formulated as a Markov Decision Process (MDP) defined by the tuple $(\mathcal{S}, \mathcal{A}, f, r, \gamma, \mathbf{s}_0)$, where $\mathcal{S}$ is the state space, $\mathcal{A}$ the action space, $f: \mathcal{S} \times \mathcal{A} \rightarrow \mathcal{S}$ is the transition model or dynamics of the system, $r: \mathcal{S} \times \mathcal{A} \rightarrow \mathbb{R}$ the reward function, $\gamma \in (0,1)$ the discount factor, and $\mathbf{s}_0$ the initial state distribution. The goal of an RL approach is to then find an optimal policy $\pi^*$ that maximizes an agent's performance in this setting. We can define the performance of a policy $\pi$ over a fixed horizon $H$ subject to the dynamics $\mathbf{s}_{t+1} \sim f(\mathbf{s}_{t}, \mathbf{u}_{t})$ as the expected sum of discounted rewards over the horizon as
\begin{equation}
    J(\pi, f) = \mathbb{E}_{u_t \sim \pi} \left[ \sum_{t=0}^H \gamma^t r(\mathbf{s}_t, \mathbf{u}_t) \mid \mathbf{s}_0 \right].
\end{equation}
Our desired optimal policy $\pi^*$ is then the result of the optimization problem 
\begin{equation}
\pi^* = \arg\max_{\pi} J(\pi, f).
\end{equation}
In practice, however, we often lack a model of the system dynamics $f$. Consequently, in \textit{model-based} RL, we first learn an approximate model $\hat{f}$ of the dynamics from real-world data and then leverage this learned model to derive control policies that maximize our objective using an RL scheme.

\section{Experiment Platform and Setup}\label{frames}
\begin{figure}[h!]
    \centering
    \begin{subfigure}[c]{0.48\linewidth}   
        \centering
        \includegraphics[width=\linewidth]{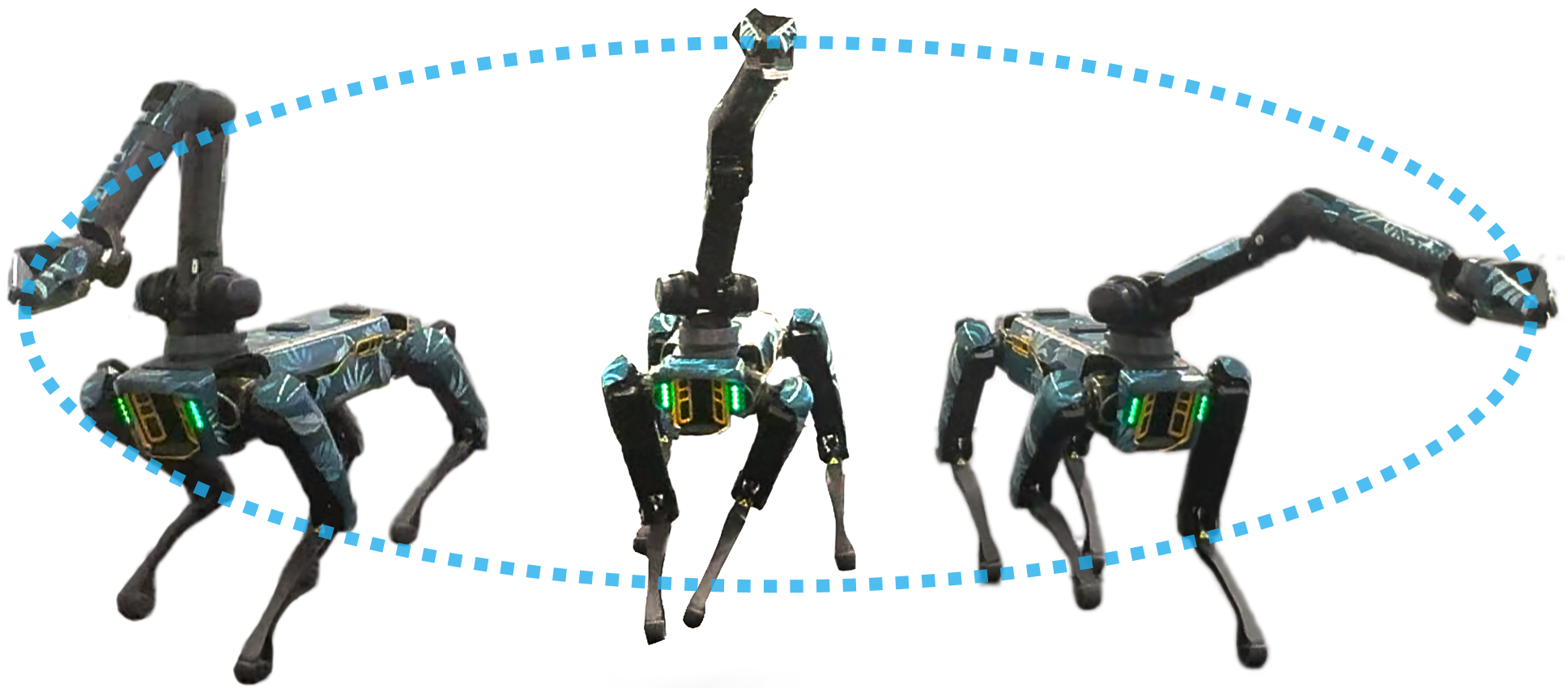}
        \caption{}
        \label{fig:ellipse_teaser_appendix}
    \end{subfigure}
    \hfill
    \begin{subfigure}[c]{0.48\linewidth}   
        \centering
        \includegraphics[width=\linewidth]{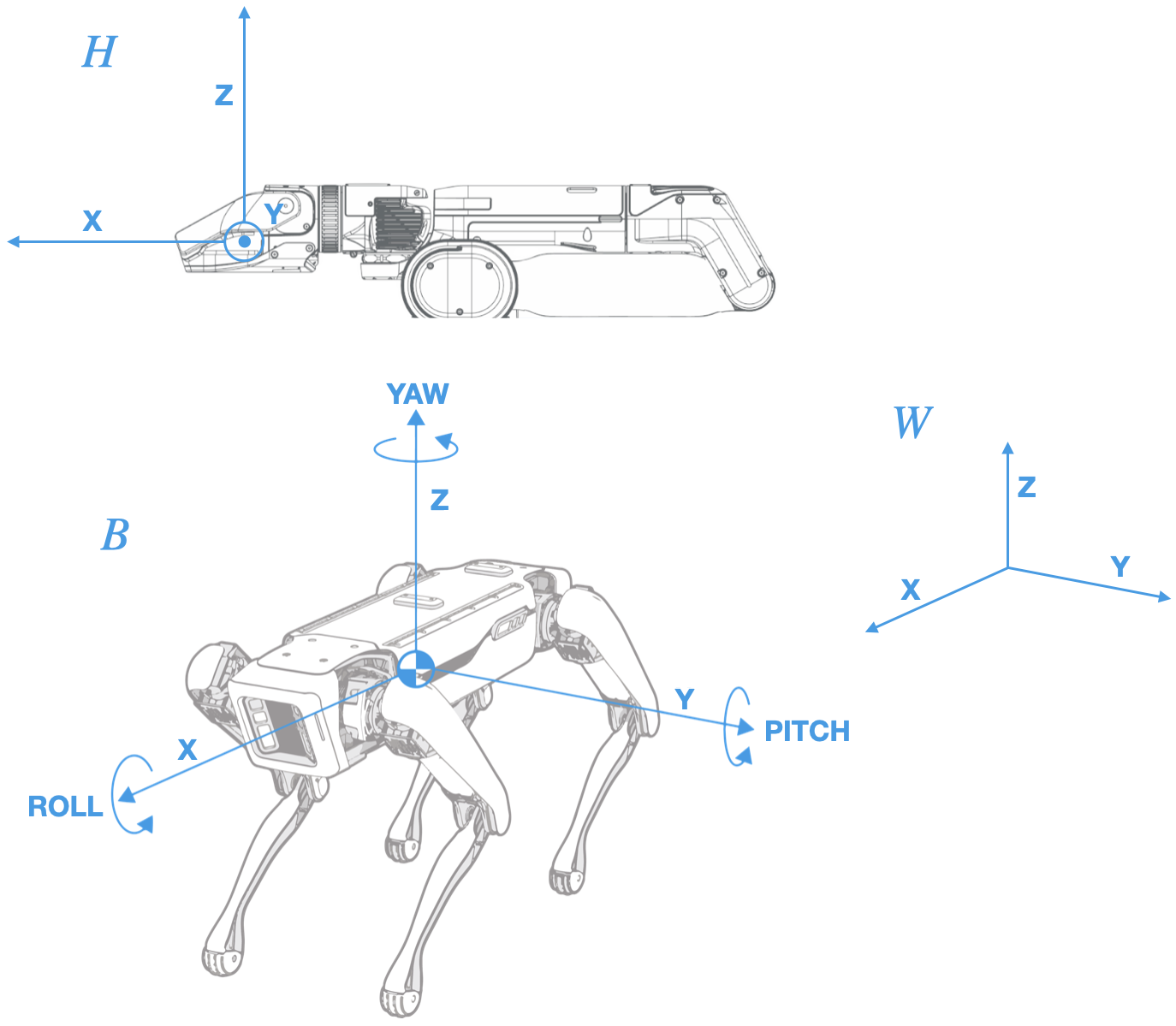}
        \caption{}
        \label{fig:spot_frames}
    \end{subfigure}
    \caption{(a) Spot following an ellipsoidal reference trajectory, and (b) the platform with the defined reference frames: the end-effector frame $H$, the body frame $B$, and the world frame $W$.}
    \label{fig:spot_combined}
\end{figure}

In \Cref{fig:spot_combined}, we show Boston Dynamics Spot following an ellipsoidal reference trajectory in our experiments, as well as an overview of the platform's geometry and its reference frames.



\section{Shape Tracking Experiments}\label{realized_trajs}
\begin{figure}[!h]
    \centering
    \vspace{0.2cm}
    \includegraphics[width=1.0\linewidth]{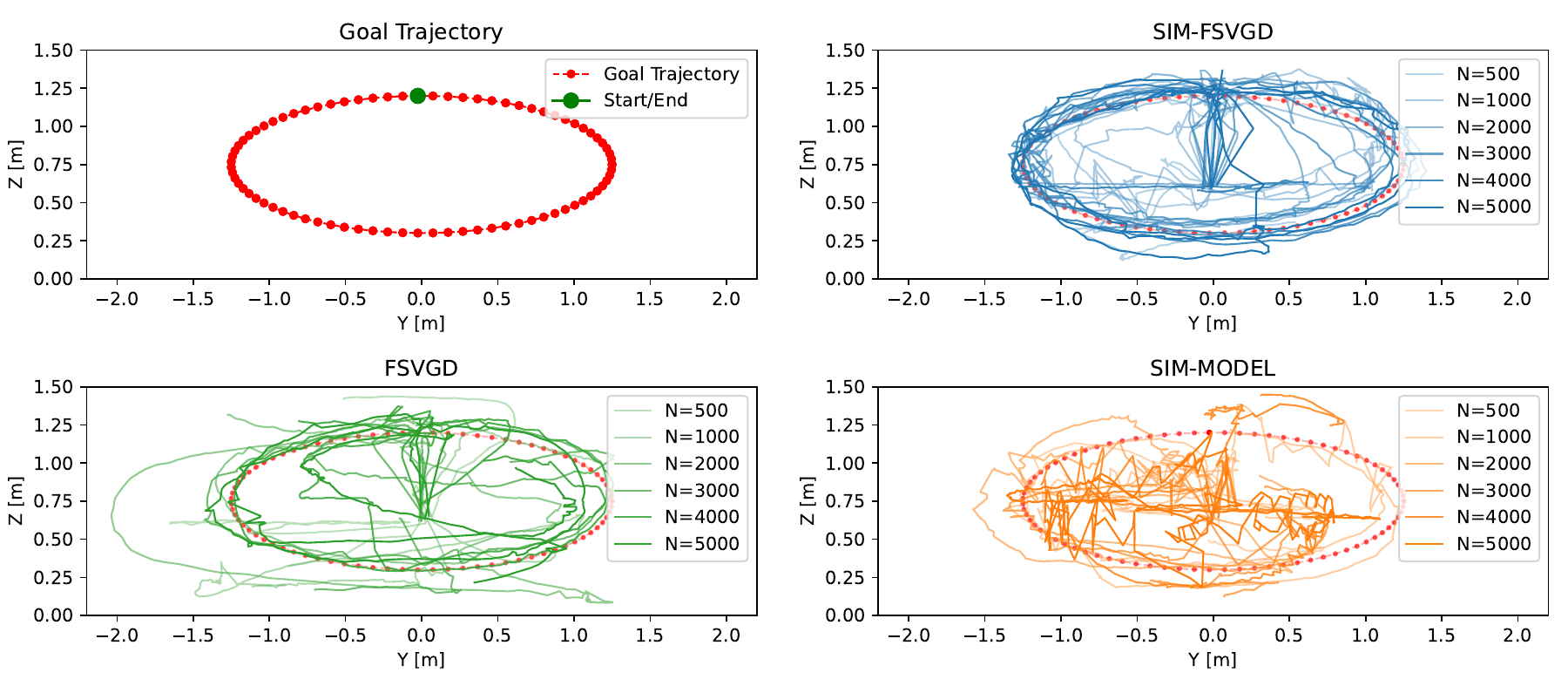}
    \caption{We plot the realized ellipse trajectories for all sample sizes and seeds. The policies learned using the {\scshape Sim}-FSVGD model follow the reference trajectory more closely than both baselines. Note that trajectories on course for collision were stopped early, in which case we plot only a truncated version. Also, the initial position of the end-effector and the first goal are not equivalent, resulting in the lines connecting the center to the first goal.}
    \label{fig:ellipse_traj}
\end{figure}

In \Cref{fig:ellipse_traj}, we provide supplementary plots that show the reference and realized trajectories for the ellipse shape tracking experiments for all polices, sample sizes and seeds.

\section{Model and Policy Learning Hyperparameters}  \label{hyperparameters}
We provide the hyperparameters we used during model and policy learning in \Cref{tab:hyperparameters_table}.
\clearpage
\begin{table}
\centering
\begingroup
\setlength{\tabcolsep}{1pt}
\begin{minipage}[t]{0.45\textwidth}
\centering
\begin{tabular}[t]{ll}
\toprule
\textbf{General BNN Hyperparameters} &  \\
\midrule
Particles & 5 \\
Batch size & 64 \\
Epochs & 100 \\
Max. training steps & 200'000 \\
Learning rate & 1e-3 \\
Weight decay & 1e-3 \\
Hidden layer sizes & 64, 64, 64 \\
Hidden activation function & LeakyReLU \\
Learn likelihood std & Yes \\
Likelihood exponent & 1.0 \\
Predict state difference & Yes \\
\bottomrule
\\
\toprule
\textbf{FSVGD Hyperparameters} &  \\
\midrule
Bandwidth SVGD & 5.0 \\
Lengthscale GP prior & 0.2 \\
Outputscale GP prior & 1.0 \\
Measurement points & 16 \\
\bottomrule
\\
\toprule
\textbf{{\scshape Sim}-FSVGD Hyperparameters} &  \\
\midrule
Bandwidth SVGD & 5.0 \\
Lengthscale physical prior & 1.0 \\
Outputscale physical prior & 0.2 \\
Measurement points & 64 \\
Function samples & 256 \\
Score estimator & GP \\
\bottomrule
\\
\toprule
\textbf{{\scshape Sim}-MODEL Hyperparameters} &  \\
\midrule
Optimizer & Adam \cite{kingma_adam_2017} \\
Training steps & 10'000 \\
Learning rate & 1e-3 \\
Weight decay & 1e-3 \\
\bottomrule
\end{tabular}
\end{minipage}%
\hspace{0.08\textwidth}
\begin{minipage}[t]{0.45\textwidth}
\centering
\begin{tabular}[t]{ll}
\toprule
\textbf{SAC Hyperparameters} &  \\
\midrule
Environment steps & 2'500'000 \\
Episode length & 120 \\
Action repeat & 1 \\
Environment steps between updates & 16 \\
Environments & 64 \\
Evaluation environments & 128 \\
Learning rate $\alpha$ & 1e-4 \\
Learning rate policy & 1e-4 \\
Learning rate q & 1e-4 \\
Weight decay $\alpha$ & 0.0 \\
Weight decay policy & 0.0 \\
Weight decay q & 0.0 \\
Max. gradient norm & 100 \\
Discounting & 0.99 \\
Batch size & 64 \\
Evaluations & 20 \\
Reward scaling & 1.0 \\
$\tau$ & 0.005 \\
Min. replay size & 2048 \\
Max. replay size & 50'000 \\
Gradient updates per step & 1024 \\
Policy hidden layer & 64, 64 \\
Policy activation function & Swish \\
Critic hidden layer & 64, 64 \\
Critic activation function & Swish \\
\bottomrule
\\
\toprule
\textbf{Reward Hyperparameters} &  \\
\midrule
$r_{\text{state}}$ bound $b$ & 0.15 \\
$r_{\text{state}}$ margin $m$ & 1.5 \\
$r_{\text{state}}$ value $a$ at margin & 0.1 \\
$r_{\text{ee-base}}$ bound $l_\text{arm}$ & 1.3 \\
$r_{\text{ee-base}}$ margin $m$ & 1.3 \\
$r_{\text{ee-base}}$ value $a$ at margin & 0.1 \\
$r_{\text{action}}$ base action weight $\lambda_\text{base}$ & 2.0 \\
$r_{\text{action}}$ end-effector action weight $\lambda_\text{ee}$ & 0.5 \\
$r_{\text{state}}$ weight $w_1$ & 1.5 \\
$r_{\text{ee-base}}$ weight $w_2$ & 0.01 \\
$r_{\text{action}}$ weight $w_3$ & 0.1 \\
\bottomrule
\end{tabular}
\end{minipage}
\setlength{\abovecaptionskip}{10pt}
\caption{Hyperparameters for model and policy learning used in our experiments.} 
\label{tab:hyperparameters_table}
\endgroup
\end{table}

\end{document}